# Enhancing Identification of Structure Function of Academic Articles Using Contextual Information

Bowen Ma[1], Chengzhi Zhang[2,·], Yuzhuo Wang[2], Sanhong Deng[1]

[1] School of Information Management, Nanjing University, Nanjing, China, 210023
[2] Department of Information Management, Nanjing University of Science and Technology, Nanjing, China, 210094
mabowen@smail.nju.edu.cn, zhangcz@njust.edu.cn, wangyz@njust.edu.cn, sanhong@nju.edu.cn

**Abstract.** With the enrichment of literature resources, researchers are facing the growing problem of information explosion and knowledge overload. To help scholars retrieve literature and acquire knowledge successfully, clarifying the semantic structure of the content in academic literature has become the essential research question. In the research on identifying the structure function of chapters in academic articles, only a few studies used the deep learning model and explored the optimization for feature input. This limits the application, optimization potential of deep learning models for the research task. This paper took articles of the ACL conference as the corpus. We employ the traditional machine learning models and deep learning models to construct the classifiers based on various feature input. Experimental results show that (1) Compared with the chapter content, the chapter title is more conducive to identifying the structure function of academic articles. (2) Relative position is a valuable feature for building traditional models. (3) Inspired by (2), this paper further introduces contextual information into the deep learning models and achieved significant results. Meanwhile, our models show good migration ability in the open test containing 200 sampled non-training samples. We also annotated the ACL main conference papers in recent five years based on the best practice performing models and performed a time series analysis of the overall corpus. This work explores and summarizes the practical features and models for this task through multiple comparative experiments and provides a reference for related text classification tasks. Finally, we indicate the limitations and shortcomings of the current model and the direction of further optimization.

---

# 1 Introduction

With the rapid development of science and technology, the output of scientific literature shows explosive growth (Xia et al., 2017). The increasingly abundant academic resources also bring the complex problem of knowledge overload, which results in an obstacle for researchers to obtain high-relevance literature (Bollacker et al., 2002). To realize efficient and accurate knowledge retrieval, many endeavors focus on adding more semantic information to the retrieval corpus. Kafkas et al. (2015) improved retrieval effects for medical journal literature by providing specific retrieval within the scope of chapter semantics. Shahid et al. (2017) illustrated advanced retrieval based on retrieval sources with semantic structure. For example, a user needs to search for literature that contains specific algorithms appearing in the *Methodology* chapter. The semantic structure is essential for high-quality information retrieval because the knowledge elements play different roles in different structure function parts of literature. For example, the research objective information exits difference in *Introduction* and *Methodology* chapters in terms of the range of semantic expression and importance of the information conveyed. The typical model dividing the contents of articles into different structure function segments is the IMRaD, proposed earlier and widely used in academic articles.

In addition to optimizing retrieval system, the structure function of academic articles is often regarded as a critical location feature and introduced into the researches. For example, in terms of Information Extraction (Heffernan & Teufel, 2018; Teufel & Moens, 2002; Nguyen & Kan, 2007), the structure function of academic articles can provide semantic features for enhancing the extraction effect. The feature is also widely used in Bibliometrics (Hu et al., 2013; Echeverria et al., 2015), and Citation content analysis is a typical example (Ding, Liu, Guo, & Cronin, 2013; Bertin, Atanassova, Sugimoto, & Lariviere, 2016; Zhu, Turney, Lemire, & Vellino, 2015). Related studies have demonstrated that the features of citation location can well reveal the intent of the citation behavior and the specific function of the citation content (Lu et al., 2017; Voos & Dagaev, 2015). It can be seen that the structure function can provide logical features of research objects for bibliometric studies and promote the development of scientific evaluation.

This paper focuses on identifying the structure function of chapter granularity and constructing classification models using traditional machine learning methods and deep learning methods. We explore the following research questions.

*RQ1*. For the features of chapter title and chapter content, which is more suitable for identifying the structure function of chapter granularity?

*RQ2*. Can model performance be further improved by integrating the features of chapter title and content?

**RQ3**. Whether the partial information in the head and tail of the chapter is more conducive to the improvement of the model performance?

*RQ4*. Can the fusion of contextual information of the chapter provide more valuable features for the model?

*RQ5*. What kind of model is more "economical" and practical for the task of the identification of structure function? (The extension of previous RQ)

This work explores and preliminarily answers the five questions, and our main contributions are as follows.

First, for RQ1, this paper divides the text feature of a chapter into chapter title and content. It provides the baseline performance of the two aspects of features on different models for the subsequent improvements. Moreover, for RQ2, the two aspects of features are further integrated for better performance.

Second, this study focuses on the perspective of the amount of feature information to explore the optimization of the feature input of the deep learning model. The RQ3 is introduced from the thought of reducing feature input. We find that the partial information at the front and back of the chapter has the potential to improve model performance under a certain amount of information.

The RQ4 is thrown out based on increasing information amount. Specifically, we try to incorporate the contextual information of chapters to enrich the feature input of the deep learning model. Lu et al. (2018) pointed out that chapters and paragraphs in academic documents are designed to express the author's thoughts and research lineage according to a certain logic. Our research argues that the logic and the expression of structure function of chapters lie in the context of the whole article. Therefore, we view the previous and subsequent chapter of the chapter to be classified as the "contextual information," including chapter titles and contents. The chapter to be classified and its contextual information forms a semantic logic segment. We also conjecture that a longer literature logic segment can improve the model performance, so we try optimizing the model by setting different fusion windows of contextual information according to the number of previous and subsequent chapters. Through the experiments, we find that the contextual information of chapters can significantly improve the model performance.

Third, this paper answers RQ5 through the open test based on non-training corpus and comparative experiments in the discussion chapter. We find that deep learning models outperform traditional machine learning models in terms of both the cost of constructing feature inputs and practical results. Our work can provide a valid basis for such research tasks and provide a reference for related tasks of text classification in terms of feature input and model construction.

Finally, we adopt the best model to annotate the main conference papers of ACL from 2016 to 2020. A complete analysis of the time evolution of the chapter category labels of ACL main conference papers is conducted (data from 1989 to 2020).

## 2   Related Works

At present, the endeavors of studying the structure function of academic articles are mainly divided into two aspects. Some works aim at building the classification frameworks of structure function, which are designed based on the characteristics of the target research corpus and have internal semantic logic. Other works focus on automatically dividing the literature content into different kinds of structure functions. It should be noted that various systems of structure function correspond to different classification granularity.

## 2.1 Classification System of Structure Function of Academic Article

The classification systems of structure function can be summarized from three aspects: the annotation granularity of corpus, the classification logic, and the fields of research and application. The representative systems are shown in Table 1.

Table 1 Major classification systems of multi-granularity structure function

| Model Type | Model Name | Annotating Granularity | Research Field |
| --- | --- | --- | --- |
| Sequenced model | IMRaD (Sollaci and Pereira, 2004) | Chapter | Research paper standard pattern |
| Models based on argumentation | Argumentative Zoning I (AZ-I) (Teufel et al., 1999) | Sentence | computational linguistics |
| | Argumentative Zoning II (AZ-II) (Teufel et al., 2009) | Clause | chemistry computational linguistics |
| Ontology-based models | Core Information about Scientific Papers (CISP) (Soldatova et al., 2007) | Sentence | General experimental research |
| | Core Scientific Concepts (CoreSCs) (Liakata et al., 2010) | Sentence | physical chemistry and biochemistry |

The IMRaD model is a classical system proposed earlier and widely used in the scientific literature (Sollaci & Pereira, 2004; Nair, P. R. R. & Nair, V. D., 2014). It divides the structure function of academic articles into four parts: Introduction, Method, Result, and Discussion. The overall structure of IMRaD is concise and clear, which illustrates the general problems in the research. As the annotation scheme is chapter-based, the obtained information on the structure function is relatively broad.

Teufel et al. (1999) proposed the Argumentative Zones (AZ) model from the perspective of argumentative logic. The model summarizes the elements in scientific articles into seven categories and focuses on describing the relationship between current work and relevant work. After that, Teufel et al. (2009) refined the AZ model in terms of research innovation and obtained AZ-II.

Some researchers introduced the idea of ontology to construct the system. Soldatova et al. (2007) proposed Core Information about Scientific Papers (CISP). Eight core concepts were summarized according to the experimental research and were used to map the research route of scientific experimental study. Subsequently, Liakata et al. (2010) added three core concepts to enrich CISP and proposed Core Scientific Concepts (CoreSCs) model.

On the whole, the IMRaD model is a coarse-grained system fitting in chapter-based annotation, while other systems have finer granularity. From the perspective of application range, the coarse-grained systems are universal and less constrained by the article type and specific field of academic articles. In contrast, the IMRaD has limited ability to parse the articles suitable for a fine-grained system. From the internal logic of the classification system, IMRaD follows the symmetric linear structure to elaborate the scientific research. The AZ system focuses on the description of structure function from the perspective of scientific argument. In contrast, the CoreSCs system parses the elements of core scientific research based on the ontology method, making the CoreSCs more comprehensive and universal.

## 2.2 Identification of Structure Function of Academic Article

There are two main methods for automatic identification of the structure function of academic articles, namely, rule-based method and learning-based method. In terms of the scope of the corpus used in research, early studies paid more attention to the abstracts. With the continuous enrichment of full-text corpus and the development of research requirements, identification based on the full text has become the mainstream of the current study. An overview of the related work is organized in Table 2.
(1) Rule-based identification method
The core of the rule-based method is mining the features related to the categories of structure function, and the features are used for matching identification or providing auxiliary judgment. Ding et al. (2013) stemmed the keywords in chapter titles, compiled a dictionary mapping to chapter categories, and optimized the results using manually constructed features. Inspired by the conclusion of the work (Ding et al., 2013), Habib et al. (2019) viewed the distribution difference of the number of citations in different chapters as a characteristic for the classification of structure function. Some other studies employed systematic methods to integrate various rules to enhance the identification ability. For example, Ahmed et al. (2020) proposed a system framework containing compelling features that are often overlooked, such as the number of citations, figures, tables, and other essential features, which improved the identification effect of previous work.

Although the rule-based method can get good results in specific scenarios, with the expansion of data scale, the refinement of classification systems and limitation of available rules make it difficult to ensure the identification effect and increase the burden of the workforce.
(2) Learning-based identification method
The mainstream idea of current research views the identification of structure function as a text classification task and trains machine learning classifiers. Teufel et al. (2002) trained 80 computer-linguistics papers and adopted the Naive Bayes model to identify the structure function based on the proposed AZ system. A variety of features, such as the position and length of sentence, TF-IDF, syntactic features, citation, formula, are integrated into the model. Kambiz et al. (2018) used Support Vector Machines (SVM), Linear regression (LR), and Naive Bayes (NB) classifiers to identify the structure function of sentence granularity. They added a voting mechanism to improve the performance further.

Some scholars adopted the sequence labeling model to explore the sequence distribution between structure functions. For example, Hirohata et al. (2008) used features including N-gram, position, and proximity information and trained SVM and Conditional Random Fields (CRF) models to identify chapter categories. The research found that the CRF performed better than SVM. Lu et al. (2018) constructed the model from 3 levels of literature information, including headers, contents, and paragraphs of chapters. The CRF model was adopted in the experiment based on chapter headers. The SVM model was trained based on vocabulary, clustering, and patterns in chapter contents and paragraphs. Asadi et al. (2019) proposed a two-level method. On the first level, Logistic Model Trees and Sequential Minimal Optimization were used to divide sentences based on semantic and lexical features preliminarily. On the second level, window fusion was introduced to comprehensively analyze sentence sequences from the first level to obtain the final results.

With the rapid development of the deep learning model, the application of neural networks opens up a new horizon for identifying the structure function. Ji et al. (2019) took SVM as the baseline model and compared the two deep learning models, text-CNN and Bert, and the results showed that Bert achieved significantly better performance.

Table 2 Related work of structure function identification

| Time | Authors | Corpus | Granularity | Methods |
|---|---|---|---|---|
| 2013 | Ding et al. | 866 research papers of JASIST from 2000 to 2011 | Chapter | Rule-based approach: Dictionary matching |
| 2019 | Habib et al. | The dataset-1 containing 320 bibliographically coupled papers, and the larger dataset-2, containing 5,000 bibliographically coupled papers | Chapter | Rule-based approach: The difference in the number of citations within chapters |
| 2020 | Ahmed et al. | 5000 papers containing 39420 sections in the computer science field | Chapter | Rule-based approach: The systematic approach incorporating a variety of characteristics |
| 2002 | Teufel et al. | 80 papers in the computational linguistics field | Sentence | Naive Bayes<br>16 kinds of characteristics |
| 2008 | Hirohata et al. | Two sets of corpora ('pure' and 'expanded') containing sentences from 51,000 abstracts | Sentence | SVM and CRF<br>3 kinds of characteristics |
| 2018 | Kambiz et al. | ART corpus including 225 papers (34,455 sentences) in the field of biochemistry annotated based on the CoreSC scheme | Sentence | SVM, Simple Logistic Regression, Bayesian Network, and Multi-layer perception<br>11 kinds of characteristics |
| 2018 | Lu et al. | 300 research articles from computer science dataset | Chapter | CRF and SVM<br>Characteristics of words, clusters, patterns<br>3 levels of classification |
| 2019 | Asadi et al. | ART corpus and DRI corpus (containing 40 papers (7859 sentences) in the field of computer graphics annotated based on a two-level scheme) | Sentence | Logistic regression and Sequential Minimal Optimization<br>14 kinds of characteristics<br>sentence-based classification and window-based classification |
| 2019 | Ji et al. | 1192 papers of JASIST | Chapter | BERT pre-trained model for obtaining features<br>SVM, Text-CNN, BERT |

(3) Summary of related work

Generally, the rule-based method gradually transforms from relying on a single feature to integrating multiple features and constructing a systematic method, and the latter achieves better results. Most of the works employ traditional machine learning methods to explore the chapter classification task. In addition to the common lexical features, researchers also enrich feature combinations through the perspectives of citations, relative position, etc. The sequence annotation model such as CRF is also well-performed in this task.

Meanwhile, we can still catch the following deficiencies or vacancies in the current studies:

First, at present, few works are applying deep learning models for exploration. Moreover, there is a lack of research to compare traditional artificial feature engineering methods with deep learning methods in depth.

Second, the learning-based approach requires the support of training data. Some of the current studies use the trigger words in chapter titles to annotate the label of the structure function. Since the coverage of chapter titles is limited, accuracy is difficult to guarantee. Some other works adopt the manual annotation approach, but the number of the labeled corpus is limited.

Third, in the current research, especially the works adopting deep learning models, there is insufficient exploration to improve feature input. In addition to the algorithmic advantages of the training models, the input of features is also crucial.

This paper focuses on the vacancies in the previous studies. In the previous work, we established an online annotating platform called CSAA (Ma, Wang, & Zhang, 2020) to assist manual annotation in constructing a large-scale training corpus. Meanwhile, two kinds of methods, including traditional machine learning models and deep learning models, were adopted to explore the features of chapter title and content in basic experiments. Subsequently, three non-semantic characteristics were introduced to improve the traditional models (Ma, Zhang, & Wang, 2020). The core contribution to this paper is to improve the feature input of the deep learning model and adaptively adjust and optimize the structure of the model. Inspired by studies of citation content analysis, this paper hopes to explore whether the contextual information can reflect the structure feature of the chapters in the same way as the contextual information of citations can reveal the semantic and pragmatic features of citation sentences? This paper set up experiments based on different features for analyzing a more cost-effective approach under this task.

# 3 Methodology

## 3.1 Framework

The overall study framework of this paper is shown in Figure 1. We selected the academic articles in the field of computational linguistics as the dataset and obtained the training corpus by manual annotation assisted by a self-built platform. The traditional machine learning and deep learning models were adopted in our experiments.

Therefore, the input data of the two models were preprocessed, respectively, and the main difference lay in whether the feature selection is performed. Chapter titles do not require the process of feature selection compared to chapter contents.

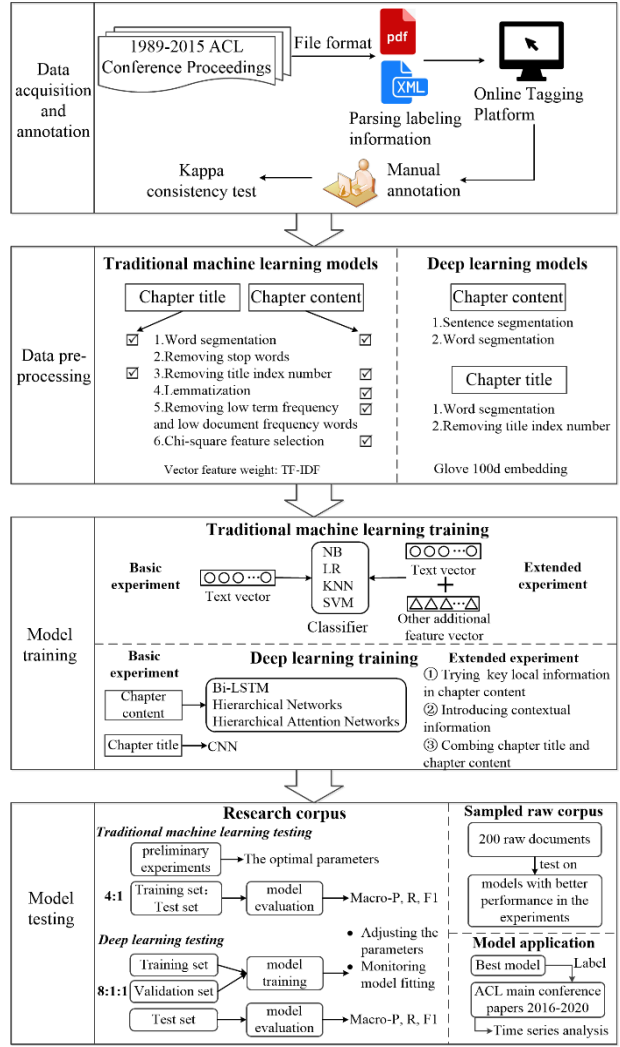

Figure 1  Framework of this study

Our experiments contain basic experiments and extended experiments. The basic experiments aim to use the lexical features of chapters to obtain the best baseline model by comparing different models. In the extended experiments, the main target is further optimizing the feature input to improve the model performance. The extension of experiments based on the traditional model lies in (1) introducing non-semantic characteristics and (2) combining the features of the chapter title and content. The extension of the deep learning model lies in (1) the partial critical information in chapter content, (2) the contextual information, and (3) combining the features of chapter title and content.

It should be noted that the model structure should be adapted to the feature input. For model testing, in addition to cross-validation of the ACL corpus used in this study, this paper also sampled 200 non-training articles from a broader document collection in computational linguistics as a raw corpus. We set the open test to verify the actual performance of our model. Furthermore, the model with the best performance in the open test was selected to label the chapter categories of the further acquired ACL 2016–2020 main conference papers. We also conducted the time series analysis based on the whole annotated corpus.

### 3.2 Data Acquisition and Annotation

This study selected the proceedings of Association for Computational Linguistics (ACL) conferences from 1989 to 2015 as corpus. ACL is the top conference in natural language processing, reflecting the newest research topics and high-quality research works, so the corpus is representative and authoritative for general research articles. The long time span of the corpus also guarantees the generalizability of the training samples. We downloaded 4190 full-text articles in XML format from the ACL Anthology Reference Corpus[1] (Bird et al., 2008). Then, the XML files were parsed and uploaded to our annotation platform with the corresponding PDF files. Annotators can refer to PDF files for convenient annotation (annotating a part of the corpus in pairs). The PDF files were downloaded from the ACL Anthology website[2].

We invited 12 undergraduate and graduate students engaged in information science research as annotators (7 graduate students and 5 undergraduates, number 3–14, divided into 6 groups). Moreover, two doctoral students as the checkers (number 1–2) review the "doubtful" tagging items given by annotators. "Doubtful" means that the annotator has particular uncertainty about his tagging result. Furthermore, we reviewed the annotation results to improve the overall accuracy.

This study used corpus in computational linguistics, most of which are research papers, including experimental parts. Therefore, we adopted the classification system proposed by Lu et al. (2018) and extended the concept they defined (1) *Method* refers to the methodology chapter, including the proposed questions, hypothesis, theoretical basis, method route, etc. So, it is not limited to the model or algorithm used in the research. (2) In addition to the experimental setup, the implementation process, evaluation method, and results analysis, this paper also included the discussion part into *evaluation & results* because we view the discussion part as an extension of the analysis of the results. For the convenience of annotation, the *other* category is set to denote the chapters that cannot be classified explicitly. Six categories of the structure function are shown in Table 3.

In the process of annotation, we found that the other chapter mainly appeared in some early articles in which the chapters were not normative. For example, in describing the methodology, some supplementary materials are introduced as separate chapters. In the newer articles, the *other* chapter mainly includes the following cases: the discussion part is included in the *conclusion* chapter, and the related work part is placed in the *introduction* or discussion part. However, such exceptional cases are fewer, so the overall effect is negligible.

---

[1] https://acl-arc.comp.nus.edu.sg/ Collection date: April, 2018
[2] https://www.aclweb.org/anthology/ Collection date: April, 2018

Table 3 The description of the structure function

| Type of Chapter | Description |
| --- | --- |
| Introduction | The research background, problems, purposes, and so on |
| Related works | A summary of the relevant work |
| Method | Description of research methods |
| Evaluation & results | Experimental setup, process, evaluation method, results, and discussion |
| Conclusion | The summary of the research and the prospects for future work |
| Other | Other chapter categories |

Before the formal annotation, we organized the tagging team to learn the annotation specification and revised it through preliminary annotation. After completing the overall task, 198 invalid articles were eliminated, including the conference overviews, reviews, etc. And then, we used the kappa coefficient to test the consistency of the annotation results (including doubtful items), as shown in Table 4.

Table 4 Statistical analysis and kappa value

| Group ID | Personnel Numbers | Article Number | Chapter Number | Kappa Value |
| --- | --- | --- | --- | --- |
| 1 | 3,4 | 657 | 3874 | 0.859 |
| 2 | 5,6 | 669 | 3998 | 0.816 |
| 3 | 7,8 | 656 | 3879 | 0.820 |
| 4 | 9,10 | 657 | 3835 | 0.726 |
| 5 | 11,12 | 693 | 4110 | 0.778 |
| 6 | 13,14 | 660 | 3896 | 0.810 |
| Overall | 3-14 | 3992 | 23592 | 0.801 |

It can be seen that the kappa coefficient of most groups is around 0.8, and the overall consistency result reaches 0.801, indicating that the annotation results are compelling. Subsequently, each group discussed the inconsistent results to reach a consensus and obtained the complete dataset for model training. The statistics of the number of chapter categories are shown in Figure 2.

From Figure 2, we can see that the number of the method is the largest because scholars set multiple chapters to introduce the methodology from multiple perspectives or at different levels. The number of the *introduction* and the *conclusion* is the same as the number of total articles essentially. There are many chapters of evaluation & result because most authors set several chapters to describe experiments, and the discussion part is usually set as a separate chapter. There is relatively little *related work* because this part is set in the *introduction* or the discussion part in some articles, and some articles do not summarize the related work in detail.

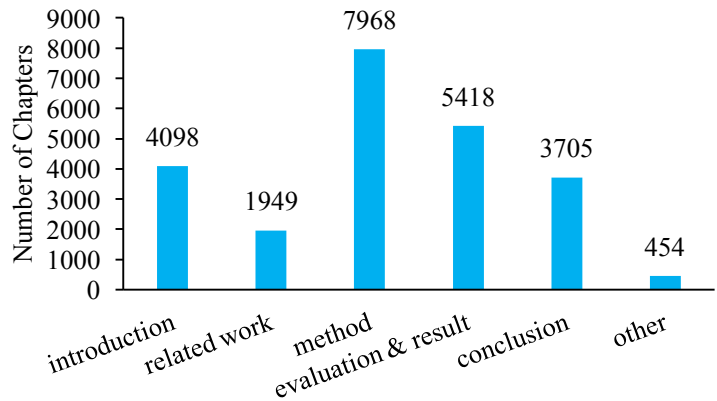

Figure 2  The number of chapters in different categories

## 3.3 The Construction of Text Feature Vectors for Traditional Machine Learning Methods

### 3.3.1 Lexical feature vector

In the experiments based on traditional machine learning models, text feature vectors were constructed according to the lexical feature of chapters. First, we performed the operations, including removing the stop words, lemmatization, and other data preprocessing operations. The dimension of the lexical feature was preliminarily reduced. Next, feature selection was further conducted. We referred to the work by Yang and Pedersen (1997), which found that information gain (IG) and the chi-square test (CHI) were the most effective methods in their experiment. In the pre-experiment, we found that CHI performed better in our task and finally obtained 5357 feature words. Then, TF-IDF was used as the feature weight to construct the text feature vector. For the lexical features in chapter titles are relatively fewer, we got 4967 feature words without feature selection, only removing the index numbers in chapter titles.

### 3.3.2 The additional non-semantic feature vector

Meanwhile, this paper also attempts to merge some non-semantic characteristics to improve the identification effect. Referring to previous works (Guo et al., 2010; Liakata et al., 2012), which focus on identifying the structure function of sentence granularity, we find that the non-semantic characteristics mainly fall into three categories: *Location*, *History*, and *Citation*. *Location* refers to the location information of the text fragment to be classified within the studied text field. *History* means the category of previous text to be classified in the same article, which conveys sequential information. *Citation* represents the number of citations in the text.

In this paper, *Location* and *Citation* were selected as additional characteristics and introduced to the classification system. *History* was excluded because the characteristic will introduce predictive information in classifying the test set, which may cause the accumulation of errors and impact the overall classification performance. We further incorporated some potential characteristics mentioned by Ahmed and Afzal (2020), which are often ignored, including the number of citations, the number of figures, the number of tables and so on. We viewed the sum of tables and figures as another characteristic explored in the experiments to simplify the additional characteristics.

It is necessary to explain that we also considered the relevant characteristics of chapter length, mainly including the number of paragraphs and sentences in the chapter. However, the original corpus we obtained has no distinct paragraph structure, and the manual correction is complex, so the characteristic is not introduced. After a preliminary experiment, it is found that the characteristic of sentence number has little effect on the results. So, the relevant characteristics of chapter length were not added. The final additional characteristics used in this paper are described below.

(1) Relative position of the chapter (Loc)

Relative position is a characteristic with potential value to the classification of structure function, the approximate interval of the chapter category can be judged. In this study, the sequence number of chapters (counting from 1) / the total number of paper chapters (except chapters with the "other" label) is taken as the value of the relative position.

(2) The number of citations (Cite)

The distribution of the number of citations usually exists differences in chapters. For example, the number of citations in the introduction and related work chapters is relatively large, while the number of citations in other chapters is small. Therefore, the number of citations (counting by the number of references) is of positive significance to the classification of structure function.

(3) Number of figures and tables (F&T)

Figures and tables are commonly used in paper writing. For experimental studies, the number of figures and tables is relatively concentrated on the experimental chapter. Therefore, we count the sum of different figures and tables in each chapter as another auxiliary characteristic.

The approach of introducing additional characteristics is to divide different value intervals for the additional characteristic and then generate a set of random numbers for each interval. Finally, the additional feature vectors are concatenated to the original vector, as shown in the equation below.

$$x_i = \left[x_i^{Lexical\ item}; x_i^{Additional\ features}\right] \quad (1)$$

Where $i$ represents the text of chapter $i$, and $x_i^{Lexical\ item}$ refers to the original text feature vector. $x_i^{Additional\ features}$ refers to the additional text feature vector, which can combine other additional characteristic vectors backward.

### 3.4 Models Adopted in Basic Experiments

This paper compared the classical machine learning models used in related studies and selected representative models for our experiments. Although the Conditional Random

Field (CRF) model is widely used, the essence of the method is to adopt the idea of sequence annotation to solve the classification problem. We will use the CRF model for comparison in the discussion chapter. In the choice of deep learning models, due to the small amount of related work available, we constructed the commonly used deep learning models based on the length of the text to be encoded. The comparison with the models used in related studies will be elaborated on in the discussion chapter.

### 3.4.1 Translated with www.DeepL.com/Translator (free version)Traditional machine learning models

Four common machine learning models were selected in this paper: Naive Bayes (NB) model (Chen et al., 2009), Logistic Regression (LR) (Lei et al., 2019), K-nearest Neighbor (KNN) (Cover & Hart, 1967), Support Vector Machine (SVM) (Cortes & Vapnik, 1995). Since our study belongs to a multi-classification task, the LR and SVM models should adopt the adaptive scheme of the task of multiple classifications. The two typical schemes are One-vs-One (OVO) (Zhang et al., 2016) and One vs. All (OVA) (Clark & Boswell, 1991). The practical method was determined by pre-experiment.

### 3.4.2 Deep learning models

For the encoding of chapter content, we chose the Bi-directional Long Short-Term Memory (Bi-LSTM) model based on time series, which is suitable for feature extraction of lengthy text. This model is a variant of Recurrent Neural Network (RNN) that learns the information on the current moment from the hidden layer through the input of different time steps. Meanwhile, the hidden layer also contains information on the previous time steps to achieve cumulative learning of the input sequence. The original Long Short-Term Memory (LSTM) model was proposed by Hochreiter et al. (1997) to optimize the status updates of the hidden layer at each time step by the "gate" structure. Bi-LSTM can encode the bidirectional sequences of input text to enhance its characterization ability of the long text sequence (Yao & Huang, 2016).

This paper referred to the Hierarchical Attention Networks (HAN) proposed by Yang et al. (2016). This model introduces the concept of hierarchy for encoding long text. In our experiment, the sentence-level hierarchical attention network was adopted, and Bi-LSTM was used as the encoding unit for each layer. The attention mechanism is to attach weight to the output of different time steps to highlight the essential features.

This study employed the Convolutional Neural Network (CNN) model to encode the short text such as chapter title, and the model was proposed by Kim et al. (2014). Different convolution kernels can extract the features in different dimensions to represent the text. The text vector matrix is constructed by word embedding.

## 3.5 The Deep Learning Models Adopted in Extended Experiments

In addition to using the classical deep learning models, this study designed and adjusted the overall framework based on the classical deep learning model for ideas we want to explore in the study.

The extended experiment based on the deep learning model should adjust the model framework according to the feature input, so the description of models is according to the different feature inputs. The core part of the extension is introducing the contextual information on chapters. The four extension ideas are as follows: (1) the fusion of contents of chapters before, current, and after, (2) the fusion of titles of chapters before, current, and after, (3) the fusion of content and title of the current chapter, (4) the fusion of contents and titles of chapters before, current and after.

(1) The fusion of contents of current and contextual chapters

In the basic experiment, the performance of Bi-LSTM, hierarchical Bi-LSTM, and hierarchical attention Bi-LSTM models was compared, and the hierarchical Bi-LSTM model performed optimally. Therefore, we selected Bi-LSTM as the feature encoding layer to design the extended model. This study referred to the Structural Bidirectional Long Short Term Memory (SLSTM) model proposed by Zhou et al. (2020) to conduct feature coding of chapter content and its contextual information, as shown in Figure 3.

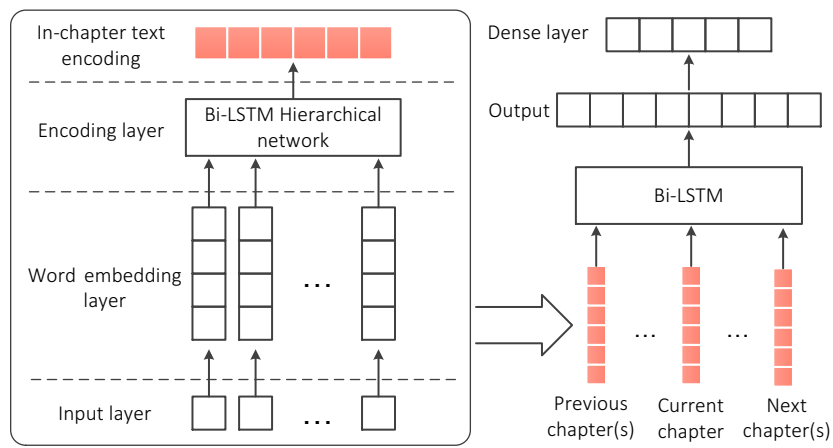

Figure 3 The model of fusing contextual in-chapter texts

Same as the basic experiments, the Bi-LSTM hierarchical model was used to encode the current chapter content. Meanwhile, the content of contextual chapters was encoded as an environmental feature. Then, according to the order of chapters in the article, we composed the initial encoding results into a new sequence of feature input and used the Bi-LSTM for the second feature encoding. Finally, the dense layer conducted the classification. The size of the fusion window determines the fusion range of contextual chapters. Suppose the current chapter does not have enough previous or subsequent chapters. The input sequence will be filled with the "padding" character, and the word vector of "padding" is a 100-dimensional zero vector.

(2) The fusion of titles of current and contextual chapters

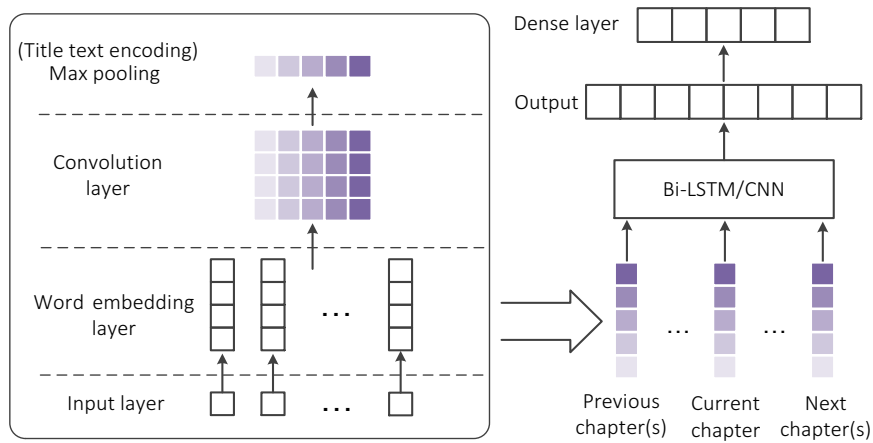

Figure 4  The model of fusing contextual chapter titles

The extended model of context based on chapter titles is similar to that based on chapter contents. We employed the CNN model to encode the feature information of current and contextual chapter titles. The Bi-LSTM was still used for integrating feature information. Moreover, considering the basic encoding unit was CNN and the consistency of pre-and post-encoding models, we also attempted to use CNN as the integration model and compared the classification effect of the two schemes. The model diagram is shown in Figure 4.

(3) The fusion of content and title of the current chapter

The experiment further tries to combine chapter titles and contents to see whether the model performance will be improved. Specifically, Bi-LSTM hierarchical model and CNN models were respectively used to encode the title and content of the current chapter. Then, two feature vectors were concatenated together to generate the overall vector.

(4) The fusion of titles and contents of current and contextual chapters

For (3), we can also introduce the concept of contextual information. Considering the feature vector of (3) is the concatenation of two aspects of chapter text features, so the sequential relationship of the text vectors is broken. It would be unreasonable to adopt the Bi-LSTM as the integration model. We thus selected the CNN model, and the feature vectors of current and contextual chapters were combined as feature matrices to enter the CNN, as shown in Figure 5.

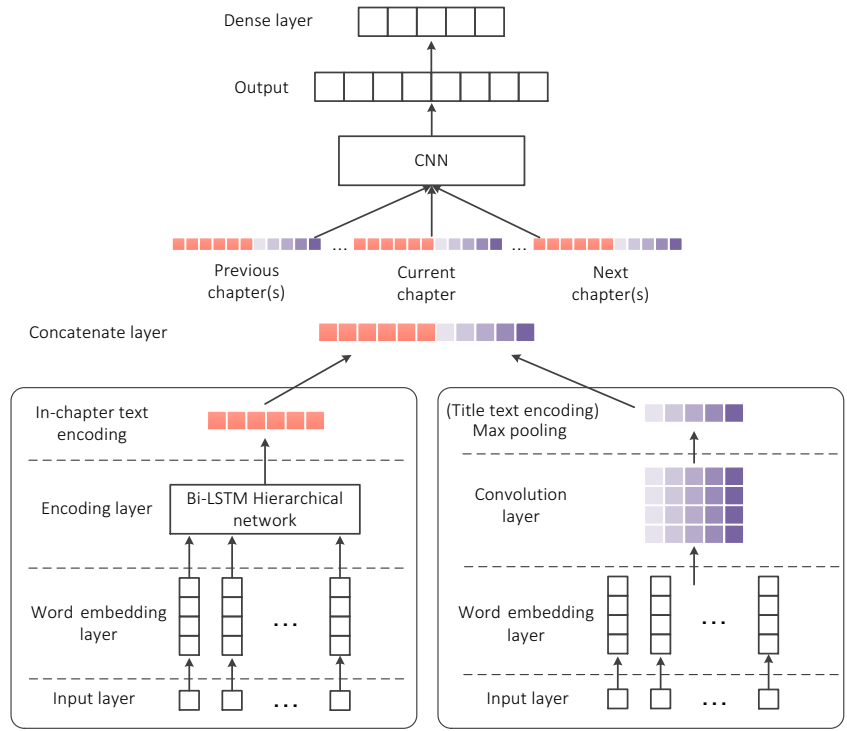

Figure 5  The model of fusing the contextual chapter titles and in-chapter texts

## 4    Experiments and Analysis of Results

### 4.1    Experimental Setup

The experiments were divided into two parts: the first adopted traditional machine learning models, and the second employed deep learning models. Each part consisted of basic experiments and corresponding extended experiments. The basic experiments mainly explore the features of chapter titles and contents, while the extended experiments further optimize the feature input. The details of the experiment are as follows:

(1) The experiments based on traditional machine learning models

   i) Experiment ML-cont: The text vector constructed based on chapter content is viewed as the feature input, and NB, LR, KNN, and SVM classifiers are selected as training models.

   ii) Experiment ML-title: The text vector constructed based on chapter title is viewed as the feature input for model training.

   iii) Experiment ML-cont-ext: Splicing three additional characteristic vectors on the text vectors based on chapter content to enrich the feature representation of chapters.

iv) Experiment ML-title-ext: Splicing the text vector of chapter content on that based on the chapter title, namely, combining two aspects of feature information. Furthermore, three additional features are integrated.

We evaluated the models of the above experiments through five-fold cross-validation. After pre-experiment, the specific parameter settings are shown in Table 5.

Table 5 The parameter settings of the traditional model

| Classifier | Parameter Settings |
|---|---|
| KNN | k =7 |
| LR | Penalty="L2", C=1.0 |
|  | adaptive scheme for multiple classifications ="One vs. One" |
| SVM | Kernel = "linear", C=1.0 |
|  | adaptive scheme for multiple classifications ="One vs. One" |

(2) The experiments based on deep learning models

i) Experiment DL-cont: Bi-LSTM, hierarchical Bi-LSTM (sentence-level), hierarchical attention Bi-LSTM model are used for feature extraction of chapter content, and then the softmax layer is used for classification.

ii) Experiment DL-title: Due to the short title of the chapter is not a complete sentence structure, this paper decides to employ the CNN model to extract feature information and then uses the softmax layer for classification.

iii) Experiment DL-cont-ext1: This paper assumes that the partial information of chapter content is more conducive to identifying the structure function, such as the front and back of the chapter. Based on this, Experiment DL-cont-ext1 extracts the text from the front and back of the chapter according to a certain proportion to construct the partial chapter information. For example, the first 10% and the last 10% of the chapter content are encoded by Bi-LSTM, respectively. Then, the two feature vectors are concatenated and accessed into the fully connected layer for classification. We further tried 20%, 30%, and 40% of the sampling ratio as comparative experiments. To further verify that the partial information at the front and back of the chapter contains more key classification features, we set up a control experiment that maintains the same amount of text as the experiment with the two parts of information. Specifically, the first 20%, 40%, 60%, and 80% of the chapter content are set as the comparative experiment group. In summary, this extended experiment explores the construction of high-quality text features to identify structure function from the two perspectives of information content and partial critical information.

iv) Experiment DL-cont-ext2: If we regard a chapter as a basic unit, the previous and subsequent chapter(s) constitute its context environment. This study hypothesizes that the context environment can provide boundary information between chapters for the classification system, which can also be understood as serialized contextual information. So this paper design Experiment DL-cont-ext2, which encodes the content of the current chapter and the surrounding chapters. The feature vectors are merged as the representation of the current chapter. Considering the number of chapters in general research articles, we design three groups of experiments that integrate content information of 1, 2, or 3 chapters previous and subsequent. In each group of experiments, the current chapter is set as $S$; the previous chapter content is set as

$S_{previous}$; the subsequent chapter content is set as $S_{next}$. In addition, we set three kinds of feature input modes for comparison, including $S_{previous} + S$, $S+S_{next}$, $S_{previous} + S+S_{next}$.

v) Experiment DL-title-ext1: Similarly, for chapter titles, contextual information can also be introduced. The specific process is the same as the Experiment DL-cont-ext2.

vi) Experiment DL-title-ext2: Splicing the feature vectors of chapter title and content to combine the two aspects of text feature, then viewing the combined feature vector as a basic unit and introducing the contextual information to optimize the model performance further.

In the above experiments, the corpus was divided into the training set, test set, and validation set according to 8:1:1. The specific parameter settings of the model are shown in Table 6.

Table 6 The parameter settings of the deep learning model

| Parameter | Value |
| --- | --- |
| Learning rate | 0.01 |
| Batch size | 128 |
| Dropout | 0.5 |
| Bi-LSTM hidden size | 100 |
| CNN filter number | 50 |
| CNN filter height | 1, 2, 3 |

(3) The test of the non-training corpus of ACL Anthology

To verify the mobility and applicability of our model, we tested the model in a broader range of non-training corpus. Specifically, we took all the articles on the ACL Anthology website as the sample space (excluding the articles used in the model training, namely, the proceedings of the ACL conference) and randomly selected 200 articles for the model verification. Two previous annotators were invited to annotate the sampling corpus. After passing the consistency test (kappa coefficient is 0.825), the remaining annotation items were discussed and unified by annotators, and 1915 pieces of test data were finally obtained. This paper chose the model with better performance in the previous experiments to conduct the test. The traditional models were retrained on the whole data sets, and the trained deep learning models were directly used because of the high time cost of retraining.

## 4.2  Evaluation Method

In this paper, the precision (P), recall (R), and $F_1$-score ($F_1$) values were used to evaluate the classification performance of each model. The specific calculation formulas are as follows.

$$P = \frac{The\ number\ of\ chapters\ correctly\ classified\ as\ class\ C}{The\ total\ number\ of\ chapters\ annotated\ as\ class\ C\ by\ classifier} \quad (2)$$

$$R = \frac{The\ number\ of\ chapters\ correctly\ classified\ as\ class\ C}{The\ total\ number\ of\ chapters\ annotated\ as\ class\ C\ by\ human} \quad (3)$$

$$F_1 = \frac{2 \times P \times R}{P + R} \quad (4)$$

Where C is a certain category of the chapter. To measure the performance of the overall classification system, we used the macro-average of precision, recall, and $F_1$-score values as the evaluation indexes. The specific calculation formulas are as follows.

$$Macro\_P = \frac{1}{n}\sum_{i=1}^{n} P_i \quad (5)$$

$$Macro\_R = \frac{1}{n}\sum_{i=1}^{n} R_i \quad (6)$$

$$Macro\_F_1 = \frac{2 \times Macro\_P \times Macro\_R}{Macro\_P + Macro\_R} \quad (7)$$

Where $n$ is the number of classification categories. $P_i$ and $R_i$ represent the precision and recall value of a specific category of chapters.

## 4.3 Results

### 4.3.1 The results based on traditional machine learning models

(1) The results of basic experiments

**We answer RQ1 in this subsection from the view of traditional machine learning models.** The results of the basic experiment based on the traditional machine learning model are shown in Table 7.

Table 7 Results of Experiments ML-cont and ML-title

|  | Model | Macro-P | Macro-R | Macro-$F_1$ |
|---|---|---|---|---|
| Experiment ML-cont (chapter content) | NB | 0.7324 | 0.5847 | 0.5979 |
|  | LR | 0.7799 | 0.7258 | 0.7436 |
|  | KNN | 0.3618 | 0.3200 | 0.3035 |
|  | SVM | 0.7772 | 0.7346 | **0.7501** |
| Experiment ML-title (chapter title) | NB | 0.9324 | 0.9092 | 0.9187 |
|  | LR | 0.9426 | 0.9128 | **0.9249** |
|  | KNN | 0.9197 | 0.8981 | 0.9074 |
|  | SVM | 0.9407 | 0.9138 | **0.9249** |

From the results of experiment ML-cont in Table 7, it can be seen that the performance of LR and SVM are obviously better than NB and KNN, and the $F_1$ scores of the two models are essentially the same. KNN performs the worst, and the overall classification performance of the model based on chapter content is lacking. In contrast, the model based on chapter title is significantly better, with the $F_1$ scores of all classifiers above 0.9. The LR and SVM models are also outstanding, but the difference between the classifiers is much smaller than that in experiment ML-cont.

(2) The results of extended experiments
**We answer RQ2 in this subsection from the view of traditional machine learning models.** The results of the extended experiment based on traditional machine learning models are shown in Table 8. The feature extension approach of chapter content refers to the fusion of the three non-semantic characteristics, including the relative position of the chapter, the number of citations, and the number of figures and tables in the chapter. The extension to chapter title features is the fusion of chapter content information. Further, the additional non-semantic characteristics can be attached to the feature vector to optimize the model.

From the result of experiment ML-cont-ext in Table 8, it can be found out that all models are greatly improved after integrating additional characteristics. NB and KNN are more evident among the models, and the $F_1$ scores reach more than 0.75, and $F_1$ scores of LR and SVM are further improved to 0.84. The above results include whole three non-semantic characteristics, and it should be considered that the impact of characteristics on the model may be harmful. To further explore the impact of three additional characteristics on the model performance, this paper took the LR model with the highest $F_1$ score as an example to analyze the impact of different characteristic combinations on the effect.

We experimented that all characteristics can improve the model performance, and the differences lie in the degree of improvement. The relative location (*loc*) of the chapter plays the most critical role. Only by adding the *loc* characteristic, the result can approach the one obtained by adding all three characteristics. The effect of *cite* for performance is relatively small, while that of *f&t* is not obvious.

From the experiments without fusing additional characteristics in ML-title-ext, it can be found that each model is improved compared with that without fusing the chapter content feature. The LR and SVM models are still well-performed and increase the $F_1$ score to 0.93, KNN also has a more significant improvement, but NB has a negligible effect. On this basis, when the additional characteristics are further introduced, the results are improved again. The LR achieves the highest $F_1$ score (0.9422), which is the optimal experimental result based on the traditional machine learning model.

Table 8 Results of Experiments ML-cont-ext and ML-title-ext

|  | Model | Macro-P | Macro-R | Macro-$F_1$ |
|---|---|---|---|---|
| Experiment ML-cont-ext (chapter content) | NB+characteristics | 0.8377 | 0.7586 | 0.7755 |
|  | LR+ characteristics | 0.8609 | 0.8389 | **0.8476** |
|  | KNN+characteristics | 0.7842 | 0.7465 | 0.7582 |
|  | SVM+characteristics | 0.8568 | 0.8351 | 0.8439 |
|  | LR | 0.7799 | 0.7258 | 0.7436 |
|  | LR+loc | 0.8584 | 0.8226 | 0.8344 |
|  | LR+cite | 0.7845 | 0.7484 | 0.7620 |
|  | LR+f&t | 0.7800 | 0.7340 | 0.7492 |
|  | LR+loc+cite | 0.8594 | 0.8339 | 0.8433 |
|  | LR+loc+f&t | 0.8582 | 0.8291 | 0.8396 |
|  | LR+cite+f&t | 0.7831 | 0.7556 | 0.7658 |
|  | LR+loc+cite+f&t | 0.8609 | 0.8389 | **0.8476** |
| Experiment ML-title | NB | 0.9365 | 0.9076 | 0.9201 |

|  | | Macro-P | Macro-R | Macro-F₁ |
|---|---|---|---|---|
| -ext (chapter title+content) | LR | 0.9467 | 0.9210 | **0.9322** |
| | KNN | 0.9229 | 0.9052 | 0.9127 |
| | SVM | 0.9423 | 0.9223 | 0.9313 |
| | NB+characteristics | 0.9429 | 0.9143 | 0.9264 |
| | LR+characteristics | 0.9530 | 0.9330 | **0.9422** |
| | KNN+characteristics | 0.9342 | 0.9217 | 0.9272 |
| | SVM+characteristics | 0.9487 | 0.9315 | 0.9395 |

#### 4.3.2 The results based on deep learning models

(1) The results of basic experiments

**We answer RQ1 in this subsection from the view of deep learning models.** The results of basic experiments based on the deep learning model are shown in Table 9.

Table 9 Results of Experiments DL-cont and DL-title

| | Model | Macro-P | Macro-R | Macro-F₁ |
|---|---|---|---|---|
| Experiment DL-cont (chapter content) | Bi-LSTM | 0.8504 | 0.8214 | 0.8327 |
| | Bi-LSTM+ hierarchy | 0.8619 | 0.8371 | **0.8471** |
| | Bi-LSTM+hierarchy+attention | 0.8545 | 0.8267 | 0.8370 |
| Experiment DL-title (chapter title) | CNN | 0.9414 | 0.9092 | **0.9217** |

From Table 9, it can be found that the hierarchical network has a better effect than the non-hierarchical network. However, after adding the attention mechanism, the performance of the hierarchical network is reduced to the same level as the non-hierarchical network. It is speculated that after the weight highlights the importance of certain input words, it weakens the grasp of the global features of the text sequence. However, the prominent feature words are not enough to greatly improve the classification ability of the model, thus leading to the $F_1$ score falling rather than rising. Experiment DL-title further reflects the significance of chapter title in the identification of structure function. The $F_1$ score reaches 0.9217, which is equivalent to the optimal result in the traditional model.

(2) The results of extended experiments based on chapter content

**We answer RQ3 and RQ4 with the chapter content as feature input in this section.** The result of experiment DL-cont-ext1 is shown in table 10, "head" indicates a certain proportion of text from the beginning to the end of the chapter, and "tail" means a certain amount of text taken from the negative direction.

As shown in Table 10, in the condition of the same amount of information, taking the partial content at the front and back of the chapter as feature input is better than taking just the previous part of the chapter. It can be seen that using a certain amount of partial critical information at the front and back of the chapter is more effective for this task. We can find that when the amount of text information is small, the results are generally lower than that of the baseline model, and the $F_1$ score is improved to a certain extent with the increase in information input. However, the $F_1$ score of each experimental group is between 0.74 and 0.85. The overall difference is not that great, showing that the model training based on chapter content is less sensitive to the amount of information.

From another perspective, the distribution of the practical features of classification in the chapter content is dispersed, explaining that the model performance is less affected by the amount of text information. The model containing 40% partial information in front and back of the chapter is slightly better than the baseline model, showing that partial critical information can better condense valuable features based on a certain amount of information input. Nevertheless, in this study, limited by the insensitivity of the information amount, the improvement effect of this approach is relatively small.

Table 10 Results of Experiment DL-cont-ext1

| Experimental group | Macro-P | Macro-R | Macro-$F_1$ |
| --- | --- | --- | --- |
| Baseline | 0.8619 | 0.8371 | 0.8471 |
| 10% (head+tail) | 0.8036 | 0.7730 | 0.7837 |
| 20% (head+tail) | 0.8257 | 0.8099 | 0.8164 |
| 30% (head+tail) | 0.8526 | 0.8256 | 0.8367 |
| 40% (head+tail) | 0.8630 | 0.8405 | **0.8486** |
| 20% (head) | 0.7672 | 0.7346 | 0.7464 |
| 40% (head) | 0.7979 | 0.7947 | 0.7942 |
| 60% (head) | 0.8280 | 0.8209 | 0.8229 |
| 80% (head) | 0.8591 | 0.8222 | 0.8367 |

The experiment DL-cont-ext2 based on chapter content mainly introduces contextual information. To compare with the baseline result of experiment DL-cont reasonably, we used the complete chapter content to conduct the experiment. The specific result is shown in Table 11. The label "Around1" means the fusion of one previous chapter and the next chapter, and "Around2" refers to the fusion of two previous and next chapters. In the same way, "Around3" means the size of the fusion window is three. The "previous" represents the fusion of only forward chapters, "next" represents the fusion of only backward chapters, and "previous + next" means the fusion of bidirectional chapters.

From table 11, we can see that all experimental groups combined with contextual information are improved compared with baseline results, and the highest $F_1$ score is increased from 0.8471 to 0.9098. From the overall view, the integration of contextual environmental information is effective for optimizing the model.

The comparative experiments exploring the directionality of fusing contextual information show that the model integrating bidirectional information has a better effect than that fusing unidirectional information. Specifically, the result of the model including bidirectional information is not that different from that with only the *previous* chapter information. However, the result is significantly better than that with the *next* chapter.

We explored the reason by the evaluation indicators of each chapter category. It is found that the integration of backward information mainly improves the identification ability of *conclusion* (the improvement is limited) but slightly weakens the identification ability of *introduction*. The introduction of forwarding information improves the identification ability of *introduction* (with a significant improvement), and slightly improves the identification performance of the *conclusion*. Regarding *related work*, the model integrating forward information is further improved, while the chapter integrating backward information gets terrible. Presumably, the reason is that the *related work* is more related to its general forward chapter, namely, the *introduction*.

Table 11 Results of Experiment DL-cont-ext2

| Experimental group | Macro-P | Macro-R | Macro-$F_1$ |
| --- | --- | --- | --- |
| Baseline | 0.8619 | 0.8371 | 0.8471 |
| Around1 (previous+next) | 0.9132 | 0.8942 | 0.9021 |
| Around1 (previous) | 0.8912 | 0.8846 | 0.8873 |
| Around1 (next) | 0.8789 | 0.8463 | 0.8589 |
| Around2 (previous+next) | 0.9142 | 0.9027 | 0.9080 |
| Around2 (previous) | 0.9017 | 0.8921 | 0.8962 |
| Around2 (next) | 0.8792 | 0.8540 | 0.8642 |
| Around3 (previous+next) | 0.9186 | 0.9024 | **0.9098** |
| Around3 (previous) | 0.9032 | 0.8881 | 0.8951 |
| Around3 (next) | 0.8826 | 0.8518 | 0.8644 |

From the perspective of the fusion window, with the increase in window size, the growth trend in the $F_1$ scores of the experimental groups integrating bidirectional information decrease gradually. The variation trend of the $F_1$ scores of the experimental groups integrating unidirectional information is approximately the same, which illustrates that the enrichment of contextual information is conducive to the improvement of classification performance. However, with the increase in information, a marginal decline will appear, which means extending the additional information to a certain extent may reduce the classification performance.

(3) The results of extended experiments based on the chapter title

**We answer RQ4 with the chapter title as feature input in this section.** In the extended experiment 1 of experiment DL-title, the contextual title information is also introduced to enhance the model. The results are shown in Table 12.

Table 12 Results of Experiment DL-title-ext1

| Experimental group | Macro-P | Macro-R | Macro-$F_1$ |
| --- | --- | --- | --- |
| Baseline | 0.9414 | 0.9092 | 0.9217 |
| Around1 (previous+next) | 0.9510 | 0.9282 | 0.9380 |
| Around1 (previous) | 0.9500 | 0.9257 | 0.9364 |
| Around1 (next) | 0.9442 | 0.9244 | 0.9335 |
| Around2 (previous+next) | 0.9506 | 0.9343 | **0.9420** |
| Around2 (previous) | 0.9503 | 0.9290 | 0.9385 |
| Around2(next) | 0.9463 | 0.9292 | 0.9372 |
| Around3 (previous+next) | 0.9514 | 0.9292 | 0.9392 |
| Around3 (previous) | 0.9552 | 0.9301 | 0.9409 |
| Around3 (next) | 0.9450 | 0.9285 | 0.9362 |

As shown in Table 12, the model's classification performance based on chapter title is also improved after integrating the forward and backward information. The overall $F_1$ scores are between 0.93 and 0.95, which is still better than the model's results based on the chapter content, with the highest $F_1$ score reaching 0.9420. Furthermore, we can see that with the increase in fusion window size, the classification effect of the model is improved first and then decreased. We speculate that the model based on the chapter title is relatively sensitive to the amount of information. The chapter title contains less feature information, so incorporating a large amount of contextual information may interfere with learning the main features in the current chapter.

Further, considering the consistency of the pre-and post-encoding, the CNN model is also employed to replace the Bi-LSTM model for feature fusion. Due to the significant effect of bidirectional information, only the experimental groups fusing bidirectional information are set under each fusion window size. The results are shown in Table 13.

Table 13 Results of Experiment DL-title-ext1 (changing fusion model)

| Experimental group | Macro-P | Macro-R | Macro-$F_1$ |
| --- | --- | --- | --- |
| Around1 (Bi-LSTM) | 0.9510 | 0.9282 | 0.9380 |
| Around1 (CNN) | 0.9528 | 0.9307 | 0.9407 |
| Around2 (Bi-LSTM) | 0.9506 | 0.9343 | 0.9420 |
| Around2 (CNN) | 0.9556 | 0.9359 | 0.9448 |
| Around3 (Bi-LSTM) | 0.9514 | 0.9292 | 0.9392 |
| Around3 (CNN) | 0.9567 | 0.9359 | **0.9454** |

It can be seen from Table 13 that the CNN model is superior to the Bi-LSTM when fusing contextual information on chapter titles. The association between the current and surrounding chapters can be seized on a broader scale through the convolution kernel of different sizes. From the experimental groups using CNN as the fusion model, we find the classification performance is gradually improved with the increase in the fusion window size, indicating that the CNN model is more suitable for feature extraction and the fusion of short text. In this extended experiment, the highest $F_1$ score is increased to 0.9454.

**We answer RQ2 in this section from the view of deep learning models.** The above experiments train the classification model based on the chapter title and chapter content, respectively. Reflecting further, can the model's performance be improved by integrating the two aspects of features? Therefore, we designed the experiment DL-title-ext2 for exploration. The results are shown in Table 14.

As can be seen from Table 14, without the introduction of contextual information, the $F_1$ scores of the models combining chapter title and content are slightly higher than that based only on chapter title (Macro-$F_1$ 0.9217). After integrating contextual information and when the fusion window size is 1or 2, the $F_1$ scores are similar to that of control groups only based on chapter title (referring to the experiment DL-title-ext1), or even slightly lower than that. When the fusion window size was set to 3, the result is further improved compared with the control group and reaches the highest $F_1$ score of 0.9471 in our study. It can be seen that the fusion of chapter title and content performs well when the fusion window is large because sufficient information enriches the compelling features for identifying structure function to a certain extent. However, the degree of performance improvement is relatively limited. The specific reasons will be analyzed in the discussion chapter.

Table 14 Results of Experiment DL-title-ext2

| Experimental group | Macro-P | Macro-R | Macro-$F_1$ |
|---|---|---|---|
| Title+Text | 0.9342 | 0.9195 | 0.9264 |
| Title+Text (Around1) | 0.9516 | 0.9297 | 0.9394 |
| Title+Text (Around2) | 0.9530 | 0.9323 | 0.9417 |
| Title+Text (Around3) | 0.9591 | 0.9371 | **0.9471** |

### 4.3.3 The results of the non-training corpus of ACL Anthology

**We answer RQ5 in this section.** This paper adopted the models that performed well in model training and practiced an open test on a sample set of 200 ACL non-training papers. According to the experimental results based on the traditional machine learning model in subsection 4.3.1, LR and SVM classifiers were selected to test three kinds of feature input: chapter content + additional characteristics, chapter title, and chapter content + additional characteristics + chapter title. The results are shown in Table 15.

Table 15 Results of open test based on traditional models

| Model | Feature Input | Macro-P | Macro-R | Macro-$F_1$ |
|---|---|---|---|---|
| LR | Text+Characteristics | 0.8539 | 0.8343 | 0.8423 |
|  | Title | 0.9349 | 0.8938 | 0.9104 |
|  | Text+Characteristics+Title | 0.9372 | 0.9119 | **0.9236** |
| SVM | Text+Characteristics | 0.8533 | 0.8297 | 0.8397 |
|  | Title | 0.9262 | 0.8924 | 0.9063 |
|  | Text+Characteristics+Title | 0.9320 | 0.9106 | 0.9206 |

Table 15 shows that the proposed models have strong adaptability and migration in a broader range of literature in computer linguistics. The models' performance is slightly lower than that on the previous test set. LR still performs better, and the best result is achieved after combining chapter content, title, and additional characteristics.

According to the experimental results of the deep learning model in 4.3.2, the optimal models in three kinds of feature input, including chapter content, chapter title, and chapter content + title, were selected for the open test. The results are shown in Table 16.

Table 16 Results of open test based on deep learning models

| Feature Input | Macro-P | Macro-R | Macro-$F_1$ |
|---|---|---|---|
| Text (Around3) | 0.9062 | 0.8899 | 0.8976 |
| Title (Around3) | 0.9509 | 0.9160 | **0.9315** |
| Text+Title (Around3) | 0.9376 | 0.9120 | 0.9238 |

Similarly, the results of the deep learning model on the raw corpus are slightly decreased. The difference from the previous experimental results lies in the classification performance of the model based on chapter title and content is significantly decreased compared with that only based on the chapter title. Compared with the traditional model, the strategy of fusing contextual information proposed in this paper can better catch the critical features in chapter title and content. The advantages of the deep learning model are obvious when only inputting chapter title features.

In summary, our proposed model based on the features of the chapter title and the contextual information performs well in practical application. The overall cost of data processing and model training is low, showing that the model is economical and reliable. The primary purpose of the open test is to validate the practical applicability of the model built in this study. The difficulty of the test derives from two aspects: (1) The selected articles are from several academic conferences in the field of computational linguistics, so there are some differences in the writing style and structure of the argument. (2) Due to the difference in the length of papers, the length of the chapters also varies significantly, which may be the possible explanation for the decline in classification performance based on the features of chapter content and title. In general, the proposed model has good adaptability and mobility in practical application.

### 4.4 Automatic annotation practice for the model based on chapter title with contextual information

In the follow-up study, we obtained additional ACL main conference paper data from 2016-2020. We adopted the best-performing model in the open test, i.e., a deep learning model trained on the chapter titles with the fusion window of 3, to annotate the chapter categories of added papers. Finally, we obtained the chapter category annotation data of ACL main conference papers from 1989–2020 and performed quantitative analysis on the corpus by time series analysis.

First, this paper analyzed the evolution trend of the percentage of each chapter category over the years. The percentage of a kind of chapter in a certain year was calculated based on the total number of a kind of chapter in a certain year/total number of chapters in the certain year. The results are shown in Figure 6.

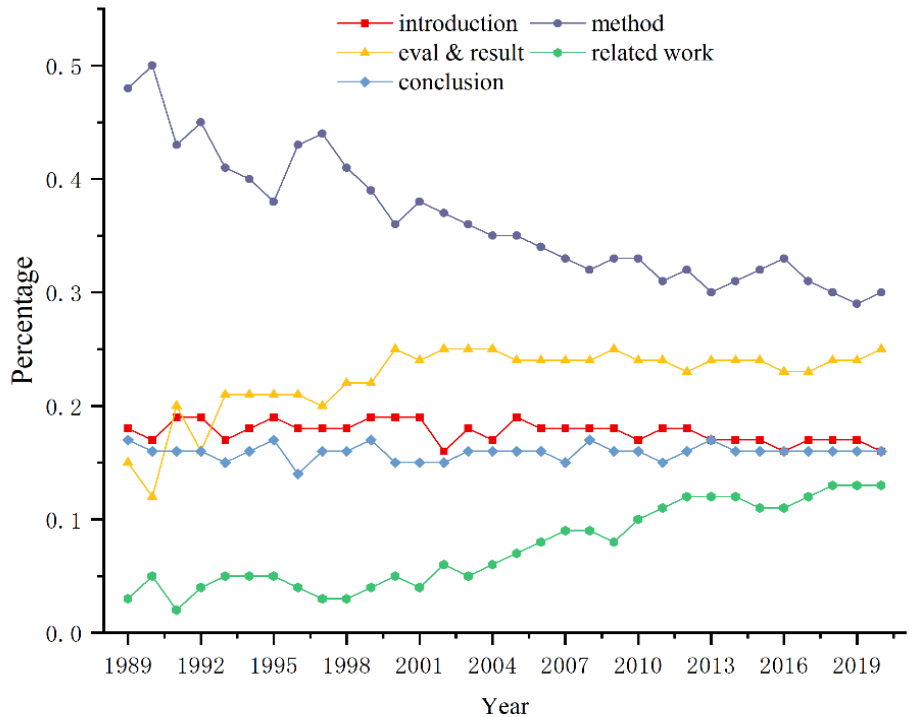

Figure 6: The time evolution figure of the proportion of each chapter category

From Figure 6, we can see that the proportion of the *introduction* and *conclusion* is stable, in line with the general structure of academic articles. It should be noted that the proportion of the *method* chapter shows an oscillating downward trend. The rate of decline is significant, and it is relatively stable after 2007. In contrast, the proportion of the *related work* shows an increasing trend, and values are gradually close to that of the *introduction* and *conclusion*. The trend indicates that the *related work* gradually becomes a fixed chapter that needs to be set independently in papers. The proportion of *evaluation & result* has increased significantly since 1990 and has stabilized since 2001. It can be found that the period of increase corresponds to a rapid decline in the proportion of the *method*. The opposite trends suggest that there is a significant increase in experiment-driven research work during this period. The subsequent period of stability also reflects that experimental research methods have become the mainstream trend in current studies. We can also infer that the standardization of paper writing is gradually improving.

Next, we counted the average frequency of different chapters in each paper over the years and drew the heat map of year evolution. The specific calculation way of the average frequency is the frequency of a kind of chapter in a certain year/the total number of papers in the certain year. The results are shown in Figure 7.

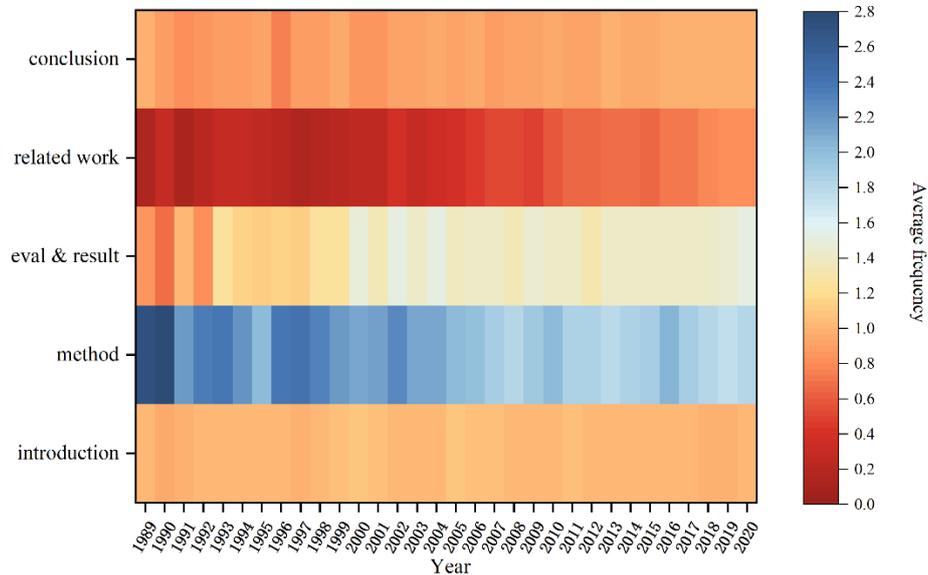

Figure 7: The time evolution figure of the proportion of each chapter category

As shown in Figure 7, the average number of the *introduction* and the *conclusion* is around 1.0 and remains stable. This paper defined the *conclusion* containing the description of future work, and some of the earlier articles divided the part of future work into individual chapters. Hence, the average frequency of the *conclusion* early on fluctuates slightly. As time evolves, the information contained in the *conclusion* chapter is relatively stable, and the number of chapters tends to be stable. The situation of *evaluation & result* is complicated. The average frequency gradually increases from below 0.8 in the early stages, exceeding the benchmark value of 1.0, and the value remains stable at around 1.4. Inferring the reason, during the labeling process of this study, we found that the chapters describing the study results generally exist in the early research literature. However, most of these studies did not adopt experimental methods. With the trend of generalization of experimental research methods, researchers generally set 1-2 chapters to describe the experimental setup and the experimental results, so the average frequency of this chapter is raised between 1.0 and 2.0. The above analysis provides a general evolution of the structure function of the overall chapters of ACL main conference papers.

## 5    Discussion

This paper analyzed the experimental results from the specific evaluation indicators of each chapter category in this chapter and compared the traditional machine learning model with the deep learning model. Moreover, we focused on the supplementary analysis of the extended experiment of the deep learning model to refine the research findings and existing deficiencies.

## 5.1 Comparative Analysis of Traditional and Deep Learning Models

(1) The comparison of basic experiments

First, we compared the basic experiments based on the traditional machine learning model with those based on the deep learning model. Specifically, The experimental groups with the best performance in experiment ML-cont-ext, ML-title, DL-cont, and DL-title were chosen to conduct comparative analysis. The results are shown in Table 17.

Compared with the models based on chapter content, the performance of the traditional models or deep learning models based on the chapter title is significantly improved, especially the *introduction* and *related work*. By comparing the optimal training results of experiment ML-cont and experiment DL-cont, the classification performance of the traditional model adding the three non-semantic characteristics in most chapters is better than that of the deep learning model except for *related work*. It shows that the *loc* feature is limited for optimizing the identification of *related work*, and the deep learning model is better for feature extraction in *related work*. Generally, the neural network is better than the traditional model in feature extraction of chapter content information, reflecting the advantage of word embedding representation to some extent. The optimal results of experiment ML-title and DL-title are equivalent. The traditional model is obviously better than the CNN in identifying *related work* and *evaluation & result*, while the identification of conclusion is worse than CNN.

Table 17 Best results of Experiment ML-cont-ext、ML-title、DL-cont、DL-title

| Chapter | Experiment ML-cont-ext | | | Experiment ML-title | | |
|---|---|---|---|---|---|---|
| | P | R | $F_1$ | P | R | $F_1$ |
| Introduction | 0.8766 | 0.9000 | 0.8881 | 0.9887 | 0.9427 | 0.9651 |
| Method | 0.8330 | 0.8919 | 0.8613 | 0.8294 | 0.9646 | 0.8919 |
| Related work | 0.7948 | 0.6400 | 0.7084 | 0.9591 | 0.9027 | 0.9300 |
| Eval & result | 0.8619 | 0.7991 | 0.8292 | 0.9384 | 0.7831 | 0.8537 |
| Conclusion | 0.9383 | 0.9636 | 0.9508 | 0.9972 | 0.9708 | 0.9838 |
| Macro average | 0.8609 | 0.8389 | 0.8476 | 0.9426 | 0.9128 | **0.9249** |
| | Experiment DL-cont | | | Experiment DL-title | | |
| Introduction | 0.8534 | 0.8805 | 0.8667 | 0.9922 | 0.9341 | 0.9623 |
| Method | 0.8110 | 0.8670 | 0.8381 | 0.8199 | 0.9711 | 0.8891 |
| Related work | 0.8726 | 0.7026 | 0.7784 | 0.9457 | 0.8923 | 0.9182 |
| Eval & result | 0.8483 | 0.7841 | 0.8150 | 0.9494 | 0.7620 | 0.8454 |
| Conclusion | 0.9239 | 0.9514 | 0.9374 | 1.000 | 0.9865 | 0.9932 |
| Macro average | 0.8619 | 0.8371 | 0.8471 | 0.9414 | 0.9092 | 0.9217 |

In order to intuitively reflect the advantages of chapter title features for the identification of structure function, the weighted CHI values of feature words in chapter title and content were calculated based on the sample proportion of each chapter category, respectively. The feature words corresponding to weighted CHI values of the top 100 were taken to plot the Pareto diagram. Due to the highest weighted CHI value of chapter title words (>4500) is far higher than other values, so we eliminated this feature word for optimizing plotting and facilitating comparison. The result is shown in Figure 8.

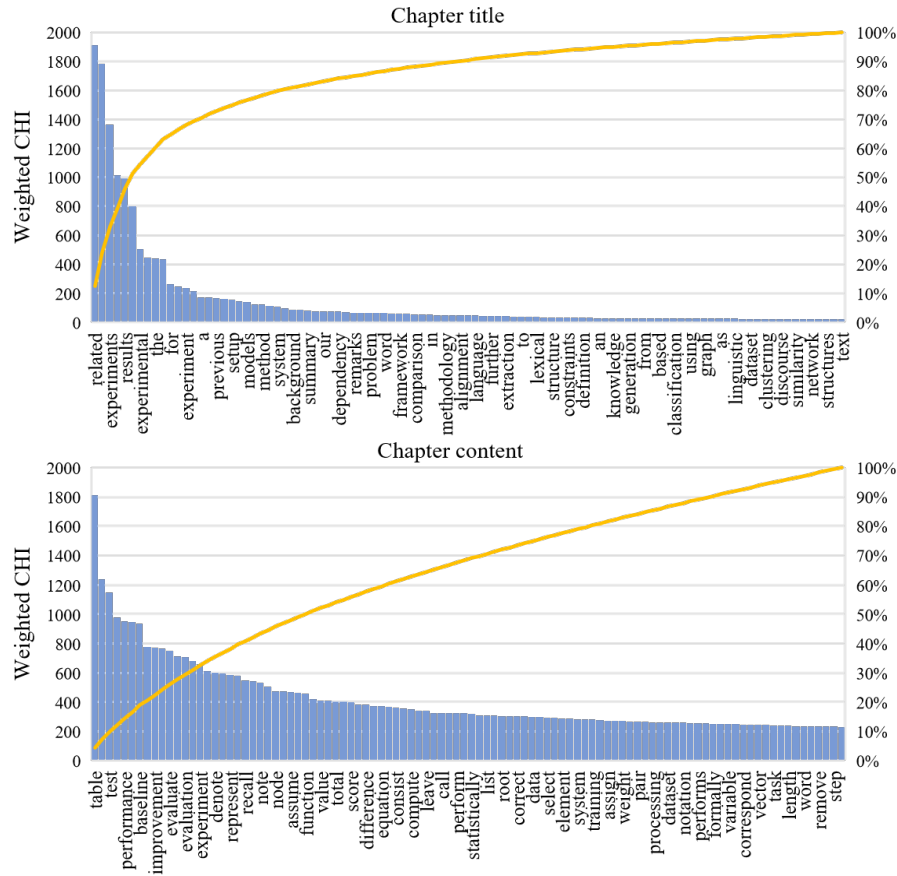

Figure 8: Comparison of weighted CHI distribution (chapter content and title)

As shown in Figure 8, the weighted CHI values of the chapter title features show an obvious polarization trend. There is a big gap between the values of high-resolution and low-resolution feature words, and the feature words with intermediate values are relatively few. However, there is no obvious polarization trend in the distribution of the chapter content features. Compared with the chapter title features, there are fewer highly differentiated features (weighted CHI value more than 1000), and the distinction between feature words is not as obvious as the chapter title features. The cumulative distribution curve can clearly reflect the difference in the distribution of discrimination of the two aspects of feature words.

(2) The comparison of extended experiments

We compared the results of the extended ML experiment with the extended DL experiment. Due to some differences in the exploring process of extended experiments ML and DL, only the experiments that fuse the features of the current chapter title and content are discussed below. So, the optimal experimental group without integrating additional characteristics in experiment ML-title-ext and the experimental group without introducing contextual information in experiment DL-title-ext2 are selected for comparison, as shown in table 18.

Table 18 Best results of Experiment ML-title-ext (without additional characteristic information) and DL-title-ext2 (without contextual information)

| Chapter | Experiment ML-title-ext | | | Experiment DL-title-ext2 | | |
| --- | --- | --- | --- | --- | --- | --- |
| | P | R | $F_1$ | P | R | $F_1$ |
| Introduction | 0.9910 | 0.9419 | 0.9658 | 0.9899 | 0.9537 | 0.9714 |
| Method | 0.8534 | 0.9615 | 0.9049 | 0.8659 | 0.9235 | 0.8937 |
| Related work | 0.9611 | 0.9000 | 0.9296 | 0.9301 | 0.8872 | 0.9081 |
| Eval & result | 0.9307 | 0.8298 | 0.8773 | 0.8878 | 0.8469 | 0.8669 |
| Conclusion | 0.9972 | 0.9716 | 0.9842 | 0.9973 | 0.9865 | 0.9918 |
| Macro average | 0.9467 | 0.9210 | **0.9322** | 0.9342 | 0.9195 | 0.9264 |

As shown in Table 18, the overall performance of the traditional machine learning model is better. From the specific indicators of each chapter category, the identification ability of other chapters is more potent than that based on the deep learning model except for the *introduction* and *conclusion*, especially for *related work*. We speculate that the training samples of related work are relatively small, limiting the ability of deep learning models to learn more valuable features. Although the traditional model has more advantages in identifying *method*, *related work,* and *evaluation & result*, the gap between the precision and recall values of the traditional model is more significant than that of the deep learning model. Therefore, the phenomenon reflects that the overall performance of the traditional model in data fitting is worse than that of the deep learning model, which will affect the generalization capability of the traditional model.

## 5.2 Analysis of Fusing Contextual Information

Next, we mainly discussed the results of the extended DL experiment integrating contextual information and explored the specific effect of fusion window size. Firstly, the experimental groups based on chapter content and contextual information in experiment DL-cont-ext2 were analyzed, as shown in Figure 9.

From Figure 9, we took the *overall* indicator as the baseline. The $F_1$ scores of *introduction* and *conclusion* are significantly higher than baseline. However, the $F_1$ scores of the remaining three chapter categories are lower than that, especially *related work*. So, the identification ability of *related work* is the most significant limitation on the overall performance. When the size of the fusion window is set to 1, the identification ability of all chapter categories is greatly improved, especially the *introduction*. We speculated that the *introduction* is placed at the beginning of the article, so its contextual environment is relatively fixed. As the size of the fusion window increases, the $F_1$ scores of most chapter categories and the baseline show a trend of moderate increase, and the increase is decreasing. However, the $F_1$ score of *conclusion* first decreased and then increased; the result is greatly affected by the window size.

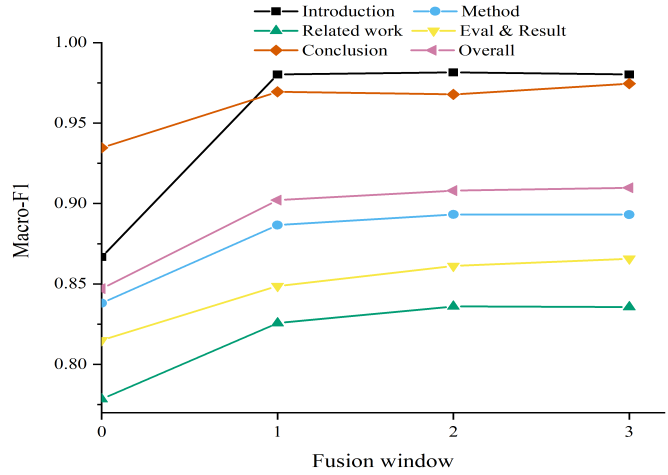

Figure 9: The figure of Macro-$F_1$ value changing with fusion window size (model based on chapter content)

The experimental group based on the chapter title in experiment DL-title-ext1 was also analyzed, as shown in Figure 10.

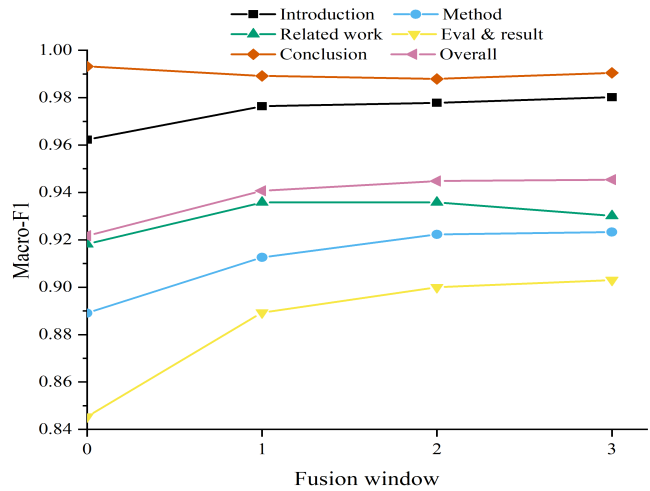

Figure 10: The figure of Macro-$F_1$ indicator changing with fusion window size (Model based on chapter title)

The general trend of the model based on chapter title is the same as the model based on chapter content, but there are some differences. First, the chapter category limiting the model's performance to the greatest extent changes from the *related work* to *evaluation & result*, indicating that the feature of chapter title is more effective for identifying the *related work*. The chapter with the highest classification accuracy also changes from *introduction* to *conclusion*. However, the classification effect of the *conclusion* decreases after the integration of contextual information. The main reason is that the *conclusion* is generally the subsequent chapter of the *evaluation and result*, but the classification performance of the *evaluation and result* is relatively poor, so the introduction of environmental information may interfere with the identification of the *conclusion*. Additionally, only the $F_1$ score of *related work* continues to decline with the increase in fusion window size, reflecting that contextual information may have a negative impact on identifying some kinds of chapters.

On the whole, the fusion of contextual environment information has an unstable influence on the chapters at the beginning or end part of the article, and the influence is considerably related to the amount of information input. For the chapters in the middle of the article, the identification ability shows an upward trend with the increase in additional information. It should be noted that improvement amplitude decreases with the increase of the fusion window size.

### 5.3 Analysis of the Limitations of Model Performance Improvement in Experiment DL-title-ext2

In the experiment DL-title-ext2, it is found that only in two situations that the $F_1$ score of the experimental group is higher than the baseline. The first is that the contextual information is not integrated, and the second is when the fusion window size is set as 3. However, the promotion is relatively limited, and we concluded two possible reasons. First, the designed model is impractical for combining the text feature of the two aspects; The second is that the chapter content cannot become the useful supplementary feature for the inadequacy of the chapter title. Considering the difficulty and high time consumption of the analysis based on the model, this paper discusses the effectiveness of chapter content and title in specific chapter categories. According to the analysis of 5.2, it can be found that three chapter categories mainly affect the overall classification performance: *method*, *related work,* and *evaluation & result*. Therefore, a comparative analysis is conducted on the evaluation indicators of these three chapters. The results are shown in Figure 11.

It can be seen from Figure 11 that the performance of the model based on chapter title is better than that of the model based on chapter content when the contextual information is not integrated, especially for the *related work* chapter. The main reason is that the chapter title of the *related work* has a precise directivity, but the chapter content features are noisy, limiting the classification effect. After introducing the contextual information, the model based on the chapter title has a certain improvement in identifying *method* and *evaluation & result*. Although the model based on chapter content has a certain improvement in identifying *method* and *evaluation & result*, the specific indicators are still lower than that of the title-based model, and the performance gap is more evident on the *related work*. Therefore, after analyzing specific indicators, we proposed the reason why the model's performance based on chapter content is lower than that based on the chapter title. Extrapolating further, even if the fusion of the two aspects of the text feature can improve the performance, the degree of improvement will be limited. Therefore, in the experiment DL-title-ext2, when the fusion window size reaches 3, the classification performance is slightly improved, showing that the extra information is still conducive to the deep learning model to capture the compelling features.

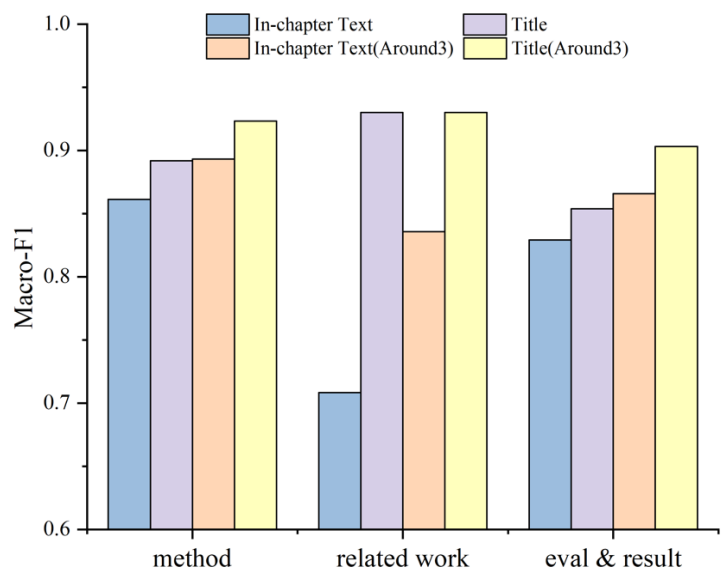

Figure 11: The comparison of Macro-$F_1$ indicator of three chapters in four groups of models

## 5.4 Analysis of the Potentials of Our Models

This subsection mainly explains the rationality and validity of our proposed model fusing contextual information in more depth.

To explain the model's plausibility and application potential, it is necessary to show that the contextual information incorporated into the feature input is valid. Therefore, we decided to illustrate this issue from two aspects: whether contextual information itself enhances the performance of the model, and the second is whether the sequential information of the chapters input into the model allows the model to obtain additional information features for learning. Then, we explored these two issues using the model that performed best in the open test as an example.

### 5.4.1 The validity of contextual information

For the first question, taking the *method* section as an example, we employed the chi-square test to calculate the chi-square values of the lexical items in the previous three chapter titles, respectively, viewing the *method* chapter as the current chapter to be classified. Through this approach, we measured the association between the contextual information and the chapter to be predicted from a side perspective of lexical items.

The results are shown in Figure 12.

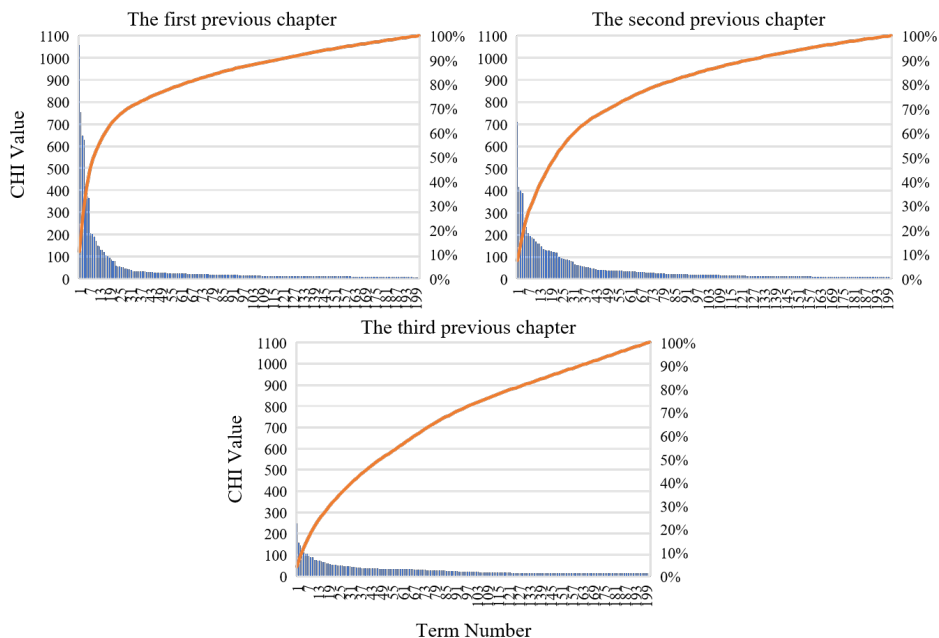

Figure 12: Chi-square distribution of the title terms of the previous three chapters of the *method* chapters

From Figure 12, the word items with Top200 CHI values were selected to plot the Pareto chart. For ease of plotting, one or two large data points in each data group were excluded. From the previous chapter of the *method*, there are more feature items with high CHI values. The cumulative distribution curve rises quickly, indicating that the presence of high-quality features in the contextual information, which are conducive to the classification. As the contextual window increases, it can be seen that the high-quality lexical items gradually decrease. From the plot of the third previous chapter, it is evident that the curve tends to flatten out. It can be inferred that the correlation between the contextual features and the current chapter is weak. The result is in line with the expectation that as the chapter distance increases, the semantic logic connection between chapters will weaken correspondingly. It should be noted that the above exploration is reflected only in lexical items. The actual model training will acquire richer semantic information, so the logical connection between chapters should be further enhanced.

### 5.4.2 The validity of chapter logical order information

For the second problem, this study attempted to change the order of inputting chapter sequences into the model to see whether the performance of the model changes. Specifically, two groups of experiments were set up: (1) The encoding information of the chapter to be classified was first input into the feature fusion layer of the model.

Then the encoding information of the previous three chapters was entered in the order of the literature. Finally, the information of the subsequent three chapters was also entered into the model. In experimental group 1, the input order of the contextual chapters was maintained, only excluding the chapter to be classified from the overall line logic. (2) The coding information of the current chapter was also input first, and then we randomly disrupted the encoding information of the contextual chapters and put the features into the model. The results are shown in Table 19.

Table 19. Results of the experimental groups controlled by the sequence of chapter feature input

| Chapter | Experimental group 1 | | | Experimental group 2 | | |
|---|---|---|---|---|---|---|
| | P | R | $F_1$ | P | R | $F_1$ |
| Introduction | 0.9925 | 0.9683 | 0.9802 | 0.9876 | 0.9683 | 0.9778 |
| Method | 0.8865 | 0.9511 | 0.9177 | 0.8639 | 0.9561 | 0.9077 |
| Related work | 0.9665 | 0.8872 | 0.9251 | 0.9763 | 0.8462 | 0.9066 |
| Eval & result | 0.9216 | 0.8672 | 0.8935 | 0.9246 | 0.8376 | 0.8790 |
| Conclusion | 0.9892 | 0.9892 | 0.9892 | 0.9865 | 0.9865 | 0.9865 |
| Macro average | 0.9513 | 0.9326 | 0.9412 | 0.9478 | 0.9189 | 0.9315 |

From Table 19, the results of both experimental groups show some decrease compared to the results in chapter 4.3.2 ($F_1$=0.9454). The indicators of experimental group 1 decrease slightly, while the results of experimental group 2 decrease to a more significant extent. It can be inferred that the input order of the contextual features has a nonnegligible impact on the model performance. In experimental group 1, only the features of the chapter to be classified are drawn out from the overall logic of the literature, so the impact on results is relatively small. However, the model performance of experimental group 2 decreases significantly. It is reasonable to believe that the randomly disordered contextual information lets the model miss the additional information coming from the complete logical sequences. Overall, the indicators of both experimental groups were higher than those of the model without introducing the contextual information ($F_1$=0.9217), which further indicates that the contextual information itself is a positive element for model training.

By analyzing the two problems, this paper believes that introducing contextual information to build a chapter classification model has certain rationality and validity and can provide references for related research.

### 5.5 The limitations of our model and the comparative discussion with related studies

In the above discussion part, this paper pointed out that our models are limited by the identification of three chapters, including *method*, *evaluation & result*, and *related work*. The best-performing model in our study still has some room for improvement in these three chapters, especially the *evaluation & result*. Meanwhile, the model's classification performance based on chapter titles is well performed, but the improvement effect is not apparent after incorporating chapter content. So the further research is needed to explore the extraction of chapter content features and the integration of content features to title features.

We hope to further analyze the advantages and limitations of our model by comparing it with the methods used in related works.

After combing through the related literature, we found no publicly available annotated datasets for the identification of structure function in chapter granularity. We referred to the two most relevant works and tested the models used in their studies on our self-built annotated dataset to perform a comparative analysis. The first is the research work of Lu Wei et al. (2018), they found that the chapter title information is direct but does not have an obvious advantage in identifying the structure function, which exists some differences compared with the results obtained in our study. The differences may be caused by the classification method we used. Lu et al. viewed the model training based on chapter title as a sequence labeling problem, and the CRF (Conditional Random Fields) (Lafferty et al., 2001) model was used for training. So we should extend our experiments for further analysis.

Specifically, we remained the features used in Lu et al.'s study to construct the CRF model for comparison (including the absolute position and relative position of the chapter title, the first two words and the last two words of the chapter title, the whole chapter title). The model was then trained on the experimental corpus and tested on the open corpus. The specific results are shown in Table 20.

Table 20. Results of the CRF model on experimental data and open data

| Chapter | Experimental data test | | | Open data test | | |
|---|---|---|---|---|---|---|
| | P | R | $F_1$ | P | R | $F_1$ |
| Introduction | 0.9897 | 0.9622 | 0.9758 | 0.9799 | 0.9420 | 0.9606 |
| Method | 0.8750 | 0.9394 | 0.9060 | 0.8667 | 0.9350 | 0.8996 |
| Related work | 0.9556 | 0.8929 | 0.9231 | 0.9412 | 0.8602 | 0.8989 |
| Eval & result | 0.8983 | 0.8508 | 0.8737 | 0.8922 | 0.8481 | 0.8696 |
| Conclusion | 0.9830 | 0.9671 | 0.9749 | 0.9774 | 0.9558 | 0.9665 |
| Macro average | 0.9403 | 0.9225 | 0.9307 | 0.9315 | 0.9082 | 0.9190 |

From Table 20, it can be seen that the CRF model performs better than the traditional machine learning models used in this study (only based on chapter title features) and can achieve an average $F_1$ value of 0.9307. However, compared with the traditional machine learning model with optimized features and the deep learning model with contextual information, the performance of the CRF model is more general, especially in the identification of *evaluation & result*. In the open test based on a non-training corpus, the actual performance of this CRF model is also worse than other models, and the performance of the model generalization is not outstanding. This comparison experiment shows that the deep learning model based on chapter titles and its contextual information has certain advantages in terms of the complexity of the feature input and the actual effect of the model.

The differences may reveal the limitation of our model caused by training data. During data annotation, the annotators relied too much on the chapter title for annotation, which leads to the chapter title features are more oriented to distinguish the chapter categories. Although in the previous training of annotation task, the annotators were required to annotate mainly according to the chapter content. It may have a tendency to rely on the chapter title in the actual tagging process, which will affect the experimental results to a certain extent.

The second is the research work of Ji et al. (2019). They compared three models SVM, CNN, and the current mainstream pre-training model Bert (Bidirectional Encoder Representation from Transformers) (Devlin et al., 2018) in the experiments. We decided to fine-turn the Bert model based on the chapter title and content information to compare with the models we constructed. The Bert model is more suitable for short text classification, and the longest input sequence length is 512. For chapter content information, only partial information can be fed into the model. To obtain a better model, three sequence input lengths of 128, 256, and 512 were tested. The batch size is set to 32 and the learning rate is set to 5e-5. For the chapter title information, the input sequence length is set to 15 and the batch size is set to 128 in previous experiments. The results are shown in Table 21.

Table 21 Results of the Bert model under different experimental groups

| Experimental group | Max sequence length | Macro-P | Macro-R | Macro-$F_1$ |
|---|---|---|---|---|
| Title | 15 | 0.9415 | 0.9188 | 0.9284 |
| Text | 128 | 0.7903 | 0.7966 | 0.7929 |
| Text | 256 | 0.8643 | 0.8308 | 0.8422 |
| Text | 512 | 0.8642 | 0.8546 | 0.8568 |
| Text (head+tail) | 512 | 0.8630 | 0.8785 | **0.8676** |

As can be seen in Table 21, the model performance based on chapter content information gradually improves as the length of the input sequence increases, and the model performs best when the input sequence is 512. Overall, the experimental results based on Bert show some improvement compared to the optimal results in Experiments DL-cont and DL-title. We further referred to the work of Sun et al (2019) and considered combining the head and tail parts of chapter content to train the model. Specifically, the first 205 characters and the last 205 characters of the chapters are coded respectively, except for the marking characters '[CLS]' and '[SEP]' in Bert. For the chapter content of fewer than 510 characters, all characters are retained and complemented with '[PAD]' characters. The results show that the model performance is significantly improved, and we can obtain a similar conclusion obtained in Experiment DL-cont-ext1, namely, more core features are distributed at the head and tail parts of chapter content.

Similarly, we tried to integrate the chapter title and content features to enhance the optimal experimental group in Table 21. The results are shown in Table 22.

Table 22 Results of extended experiment and open test based on Bert model

| Experimental group | Macro-P | Macro-R | Macro-$F_1$ |
|---|---|---|---|
| Title+Text (head+tail) | 0.9385 | 0.9291 | 0.9333 |
| Open data test | 0.9274 | 0.9160 | 0.9211 |

From Table 22, it can be seen that the model performance is further improved. Compared with the model without integrating contextual features in Experiment DL-

title-ext2, the Bert model has some advantages. However, when chapter context information is incorporated, the model based on Bert performs poorly in comparison. And the results of the open data test show that only the fine-turning of the Bert model has no significant advantage over our method in terms of training load and actual performance. Overall, the method proposed in this study is an effective improvement solution for the classical deep learning models and provides a reference for related research.

## 6 Conclusion and Future Works

In this study, large-scale full-text academic articles in the field of computational linguistics are used as the original corpus, and the training corpus of the chapter structure function is obtained through manual annotation assisted by the annotation platform. This paper trains the identification models of structure function based on traditional machine learning models and deep learning models. In the basic experiment of the traditional models, we find that the performance based on the feature of the chapter title is better than that based on the chapter content, and the LR and SVM models perform better than other models. The extended experiment in fusing three kinds of non-semantic characteristics shows that the *relative position* is a significant characteristic for identifying the structure function of chapters. After fusing the features of chapter title and content, the model performance is further improved, and the $F_1$ score could reach as high as 0.9422 when combining all three additional characteristics.

In the basic experiment based on deep learning models, we find that the Bi-LSTM hierarchical network has the most robust ability to extract the features of chapter content, which is better than the traditional machine learning models. From the extended experiment, we find that partial critical information can improve the performance of the model to some extent. However, the valuable features of the chapter content are dispersed, so the classification ability is limited when the amount of partial information is relatively small. And when the amount of content information input reaches a certain level, the advantage of partial critical features can be revealed. The satisfactory results are obtained from the exploration of chapter contextual information, and the effect of the model based on the chapter title or the chapter content is greatly optimized. With the increase in fusion window size, the improvement is continuous, but its amplitude is reduced gradually. In the end, we combine the features of the chapter title and content and observe that the traditional machine learning model has a better effect without other characteristics. After the three non-semantic characteristics are integrated into the traditional model and the contextual information is introduced into the deep learning model, the highest $F_1$ scores of both models reach above 0.94.

Moreover, in all experiments, the highest F1 score is 0.9471 in the deep learning model when the fusion window size is 3. However, considering the economics of model training, the deep learning model based on the chapter title and the contextual information can achieve a good enough performance. After sampling and conducting the open test in a broader collection of a raw corpus, we find that the evaluation indicators are equivalent to the test results of the previous model. When just adopting chapter title features, the deep learning model has obvious advantages over the traditional machine learning models and has the best classification performance. We also use the best model in the open test to annotate the ACL main conference papers in recent five years. The time series analysis based on the annotated corpus from 1989 to 2020 shows that the article structure of literature in computer linguistics is gradually stabilizing, and experimental methods are becoming the dominant research approach.

We are inspired by exploring traditional machine learning models and viewed the contextual information as the *relative position* feature to optimize the deep learning model and achieve a good result. In the future, we hope to integrate contextual information into traditional models. However, due to the high dimensional sparseness of artificial text vectors, the vector splicing is no longer applicable, so the specific fusion method needs to be further studied.

In terms of the deep learning method, it is necessary to explore how to optimize the model design to extract more targeted and practical features, especially for improving the identification ability of the *method* and *evaluation & result* chapters. For example, we can try to increase the depth of the model or construct various features into characteristic matrices. We can also introduce the concept of "channel" in the CNN model to obtain some artificially constructed high-quality features. In terms of feature selection, considering the advantages of chapter titles, the secondary or tertiary chapter titles can be further utilized to enrich the feature combination.

## Acknowledgment

This work is supported by the National Natural Science Foundation of China (Grant No.72074113) and Open Fund Project of Fujian Provincial Key Laboratory of Information Processing and Intelligent Control (Minjiang University) (No. MJUKF-IPIC201903).